\pdfoutput=1

\documentclass[11pt]{article}

\usepackage[final]{acl}

\usepackage{times}
\usepackage{latexsym}

\usepackage[T1]{fontenc}

\usepackage{times}
\usepackage{soul}
\usepackage{url}
\usepackage{amsthm}
\usepackage{booktabs}
\usepackage{algorithm}
\usepackage{algorithmic}
\usepackage{amsmath}  
\usepackage{amssymb}  
\usepackage[switch]{lineno}
\usepackage[disable]{todonotes}
\usepackage{array}
\usepackage{graphicx}
\usepackage{multirow}
\usepackage{forest}
\usepackage{xcolor}
\usepackage{makecell}
\usepackage{amssymb}
\usepackage{enumitem}

\definecolor{lightcoral}{rgb}{0.94, 0.5, 0.5}
\definecolor{lightgreen}{rgb}{0.56, 0.93, 0.56}
\definecolor{harvestgold}{rgb}{0.98, 0.85, 0.40}
\definecolor{brightlavender}{rgb}{0.75, 0.58, 0.89}
\definecolor{capri}{rgb}{0.0, 0.75, 1.0}
\definecolor{carminepink}{rgb}{0.92, 0.3, 0.26}
\definecolor{celadon}{rgb}{0.67, 0.88, 0.69}
\definecolor{darkpastelgreen}{rgb}{0.01, 0.75, 0.24}

\definecolor{hidden-draw}{RGB}{205, 44, 36}
\definecolor{hidden-blue}{RGB}{194,232,247}
\definecolor{hidden-orange}{RGB}{243,202,120}
\definecolor{hidden-yellow}{RGB}{242,244,193}
\definecolor{tree-level-1}{RGB}{245,20,85}
\definecolor{tree-level-2}{RGB}{246,86,118}
\definecolor{tree-level-3}{RGB}{248,177,193}
\definecolor{tree-leaf}{RGB}{176,230,198}

\definecolor{Self}{RGB}{255,0,128}
\definecolor{Ensemble}{RGB}{0,127,255}
\definecolor{Iterative}{RGB}{153,51,255}

\definecolor{exemplar1}{RGB}{136,98,148}
\definecolor{exemplar2}{RGB}{148,210,242}
\definecolor{knowledge1}{RGB}{249,219,152}
\definecolor{knowledge2}{RGB}{255,245,220}
\usepackage{xspace}
\newcommand{\mllm}{LLM\xspace}

\newcommand{\mycite}[1]{\citeauthor{#1}~[\citeyear{#1}]}

\usepackage[utf8]{inputenc}

\usepackage{microtype}

\usepackage{inconsolata}

\usepackage{graphicx}

\title{A Survey of Machine Unlearning in Large Language Models: \\Methods, Challenges and Future Directions}

\author{
Qing Li$^{1\dagger}$ \quad Jiahui Geng$^{1\dagger}$ \quad Herbert Woisetschläger$^3$ \quad Zongxiong Chen$^2$ \quad Fengyu Cai$^4$ \\
\textbf{Yuxia Wang}$^1$ \quad \textbf{Preslav Nakov}$^1$ \quad \textbf{Hans Arno Jacobsen}$^5$ \quad \textbf{Fakhri Karray}$^1$ \\
$^1$Mohamed bin Zayed University of Artificial Intelligence (MBZUAI), UAE \\
$^2$Fraunhofer Institute for Open Communication Systems (FOKUS), Germany \\
$^3$Technical University of Munich, Germany \\
$^4$Technical University of Darmstadt, Germany \\
$^5$University of Toronto, Canada
}

\begin{document}
\maketitle
\begin{abstract}
This study investigates the machine unlearning techniques within the context of large language models (LLMs), referred to as \textit{LLM unlearning}. LLM unlearning offers a principled approach to removing the influence of undesirable data (e.g., sensitive or illegal information) from LLMs, while preserving their overall utility without requiring full retraining. 
Despite growing research interest, there is no comprehensive survey that systematically organizes existing work and distills key insights; here, we aim to bridge this gap. We begin by introducing the definition and the paradigms of LLM unlearning, followed by a comprehensive taxonomy of existing unlearning studies. Next, we categorize current unlearning approaches, summarizing their strengths and limitations. Additionally, we review evaluation measures and benchmarks, providing a structured overview of current assessment methodologies. Finally, we outline promising directions for future research, highlighting key challenges and opportunities in the field.
\end{abstract}

\section{Introduction}

\begin{figure}[t]
    \centering
    \includegraphics[width=\linewidth]{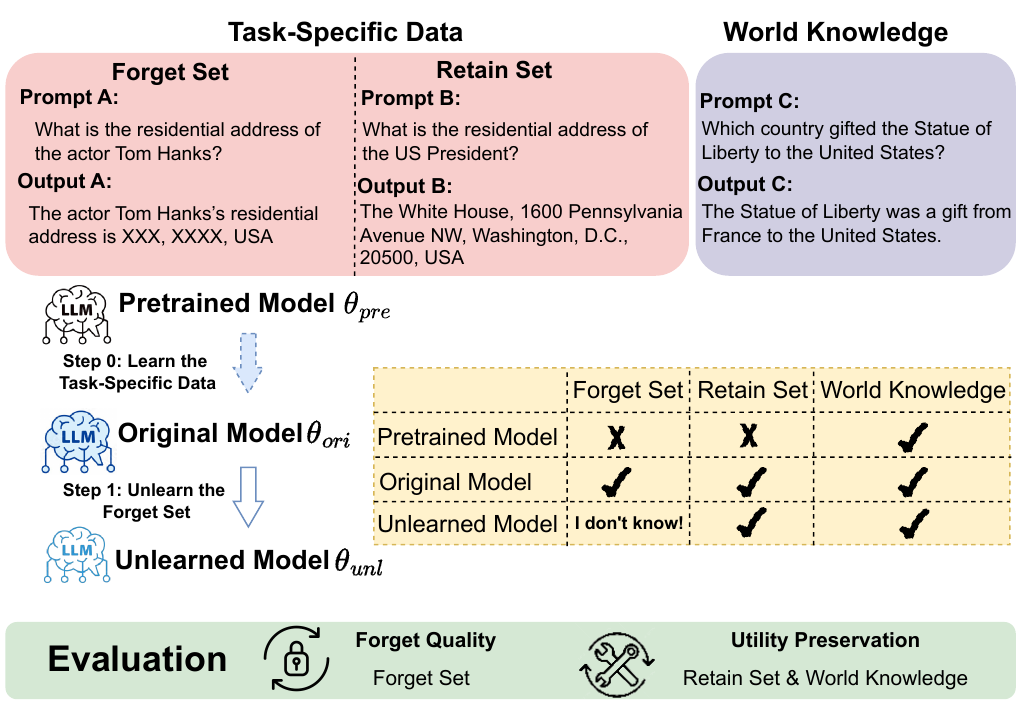}
    \caption{\textbf{Overview of LLM Unlearning.} LLM unlearning focuses on removing specific data (forget set) while minimizing the impact on related knowledge (retain set) and general world knowledge. }
    \label{fig:workflow}
\end{figure}

\begin{table*}[]
\centering
\resizebox{0.9\textwidth}{!}{
\begin{tabular}{lccccc}
\toprule
Related Work & Date & Taxonomy & \# Methodologies & Modality & Adv Evaluation  \\
\midrule
\citet{si2023knowledge}      &    December 2023          &   Yes       &      3            &  Unimodal        &             0           \\
 \citet{xu2024machine}     &       April 2024       &    Yes      &        3          &   Unimodal       &        0                \\
\citet{liu2025rethinking}      &   June 2024           &   No       &      6            &   Unimodal,Multimodal       &      5                  \\ 
\citet{liu2024machine}      &    July 2024          &   No       &        5          &      Unimodal,Multimodal    &        0                \\ \midrule
Ours      &                  May 2025      &     Yes             &     10     &  Unimodal,Multimodal  & 7      \\    \bottomrule
\end{tabular}}
\caption{Comparison of different surveys on the LLM Unlearning. \emph{Adv Evaluation} refers to Adversarial Evaluation.}
\label{tab:related_work}
\end{table*}

\tikzstyle{my-box}=[
    rectangle,
    draw=hidden-draw,
    rounded corners,
    text opacity=1,
    minimum height=1.5em,
    minimum width=5em,
    inner sep=2pt,
    align=center,
    fill opacity=.8,
]
\tikzstyle{cause_leaf}=[my-box, minimum height=1.5em,
    fill=orange!15, text=black, align=left,font=\scriptsize,
    inner xsep=2pt,
    inner ysep=4pt,
]
\tikzstyle{detect_leaf}=[my-box, minimum height=1.5em,
    fill=blue!15, text=black, align=left,font=\scriptsize,
    inner xsep=2pt,
    inner ysep=4pt,
]
\tikzstyle{mitigate_leaf}=[my-box, minimum height=1.5em,
    fill=green!15, text=black, align=left,font=\scriptsize,
    inner xsep=2pt,
    inner ysep=4pt,
]

\definecolor{rootcolor}{RGB}{138,43,226}     
\definecolor{evalcolor}{RGB}{255,140,0}      
\definecolor{methodcolor}{RGB}{30,144,255}   
\definecolor{benchcolor}{RGB}{34,139,34}     
\definecolor{appcolor}{RGB}{220,20,60}       

\forestset{
    mitigate_leaf/.style={
        fill=green!10,
        draw=green!30,
        rounded corners=2pt,
        text=black
    },
    detect_leaf/.style={
        fill=blue!10,
        draw=blue!30,
        rounded corners=2pt,
        text=black
    },
    hidden-draw/.style={
        draw=none
    },
}

The widespread adoption of large language models (LLMs) has brought significant challenges, particularly concerning user data privacy, copyright protection, and alignment with societal values. During training, these models can inadvertently memorize sensitive information, such as personally identifiable data or copyrighted materials~\cite{li2024the,li2024reference,zhang2024safe,yao-etal-2024-machine}. In addition to privacy and copyright issues, some training data may embed content that conflicts with contemporary social norms, such as discriminatory language based on race, ethnicity, etc~\cite{li2025internal}. These biases often manifest as harmful stereotypes, undermining the fairness and inclusivity of AI systems. Addressing these concerns is not only a societal imperative but also a regulatory requirement under privacy laws such as the General Data Protection Regulation (GDPR, \citealt{gdpr2022}) and the EU Artificial Intelligence Act (EU AI Act,~\citealt{com2021laying}). These laws mandate the ``right to be forgotten" and require mechanisms to delete specific data upon request.

To address these challenges, the field of LLM unlearning has emerged, focusing on removing specific information or behaviors from models while preserving their overall performance.
However, LLM unlearning faces significant technical challenges. One of the most pressing issues is the prohibitively high cost of retraining. Traditionally, addressing harmful or unwanted data required retraining the model from scratch after excluding problematic data from the training set~\cite{jang-etal-2023-knowledge}. This is impractical for LLMs due to the immense time and computational resources required~\cite{li2024the}. Moreover, the frequent unlearning requests that arise in deployed models highlight the need for more efficient unlearning techniques. The complexity of LLMs, with their millions, or even hundreds of billions of parameters, further complicates the task of removing specific information without causing unintended side effects, such as performance degradation or catastrophic forgetting~\cite{zhang2024negative}. 


Table~\ref{tab:related_work} provides a comparative overview of existing survey papers on LLM unlearning alongside our work. Most prior surveys were published before July 2024 and therefore do not capture recent advancements in the field. Moreover, these surveys either lack a systematic taxonomy or are limited to unimodal unlearning methods.
Here we aim to bridge this gap. In particular, we offer a thorough overview of the field, including various unlearning and evaluation methods, and we make the following contribution

\begin{itemize}
    \item We formalize the LLM unlearning paradigms and propose a comprehensive taxonomy to categorize the existing approaches. This taxonomy not only offers a structured understanding of the research landscape, but also helps researchers identify their areas of interest.
    \item We systematically review the existing methods, analyzing their strengths and weaknesses. We further examine the existing evaluation measures and benchmarks, highlighting the challenges of balancing utility preservation with forgetting quality.
    \item We discuss future research opportunities for LLM unlearning, including extending techniques to multimodal models and addressing complex real-world unlearning requests. These avenues aim to advance the field and address emerging challenges. 
\end{itemize}

\section{Preliminaries and Taxonomy}
\label{sec:preliminary}

\subsection{Problem Definition}

The objective of LLM unlearning is to selectively remove the influence of specific information while maintaining the model's overall utility for other tasks. The optimization objective of the model parameters $\theta$ can be expressed as follows:

\begin{equation} \min_{\theta}  \mathcal{L}(\theta) = \min_{\theta} \{ -\mathcal{L}_f(\theta) + \lambda \mathcal{L}_r(\theta) \label{eq:definition} \} \end{equation}

\noindent Here, the \textit{forget loss} $\mathcal{L}_f(\theta)$ quantifies the model prediction error on the forget set $\mathcal{D}_{f}$, while the \textit{retain loss} $\mathcal{L}_r(\theta)$ ensures the preservation of the model's utility on the retain set $\mathcal{D}_{r}$. The regularization parameter $\lambda \ge 0$ controls the trade-off between effectively forgetting undesired information and preserving the model’s utility.

\subsection{LLM Unlearning Paradigms}
LLM unlearning follows two main paradigms. The \emph{fine-tuning-then-unlearning} paradigm focuses on eliminating knowledge introduced during \textit{fine-tuning}.
As illustrated in Figure~\ref{fig:workflow}, this paradigm typically leverages synthetic data (e.g., TOFU~\cite{maini2024tofu} and FIUBench~\cite{ma2024benchmarking}) and partitions a task-specific dataset into a forget set $\mathcal{D}_f$ and a retain set $\mathcal{D}_r$ to highlight the unlearning precision. The original model \(\theta_{ori}\) is first obtained by fine-tuning a pretrained model \(\theta_{pre}\) on the task-specific data to encode the target knowledge. Unlearning techniques are then applied to reduce the model's reliance on the forget set, resulting in the unlearned model \(\theta_{unl}\).
The \emph{direct-unlearning} paradigm focuses on eliminating knowledge acquired during the \textit{pretraining} stage of $\theta_{ori}$, assuming the target knowledge originates from multiple points within the pretraining dataset. The forget and retain sets are typically sampled from pretraining corpora~\cite{yao-etal-2024-machine} or publicly available datasets such as Wiki~\cite{jin2024rwku}. This paradigm is also extensively applied in safety alignment tasks, where it aims to eradicate hazardous knowledge and to mitigate risks such as misuse or jailbreak attacks~\cite{li2024the,zhang2024safe}.

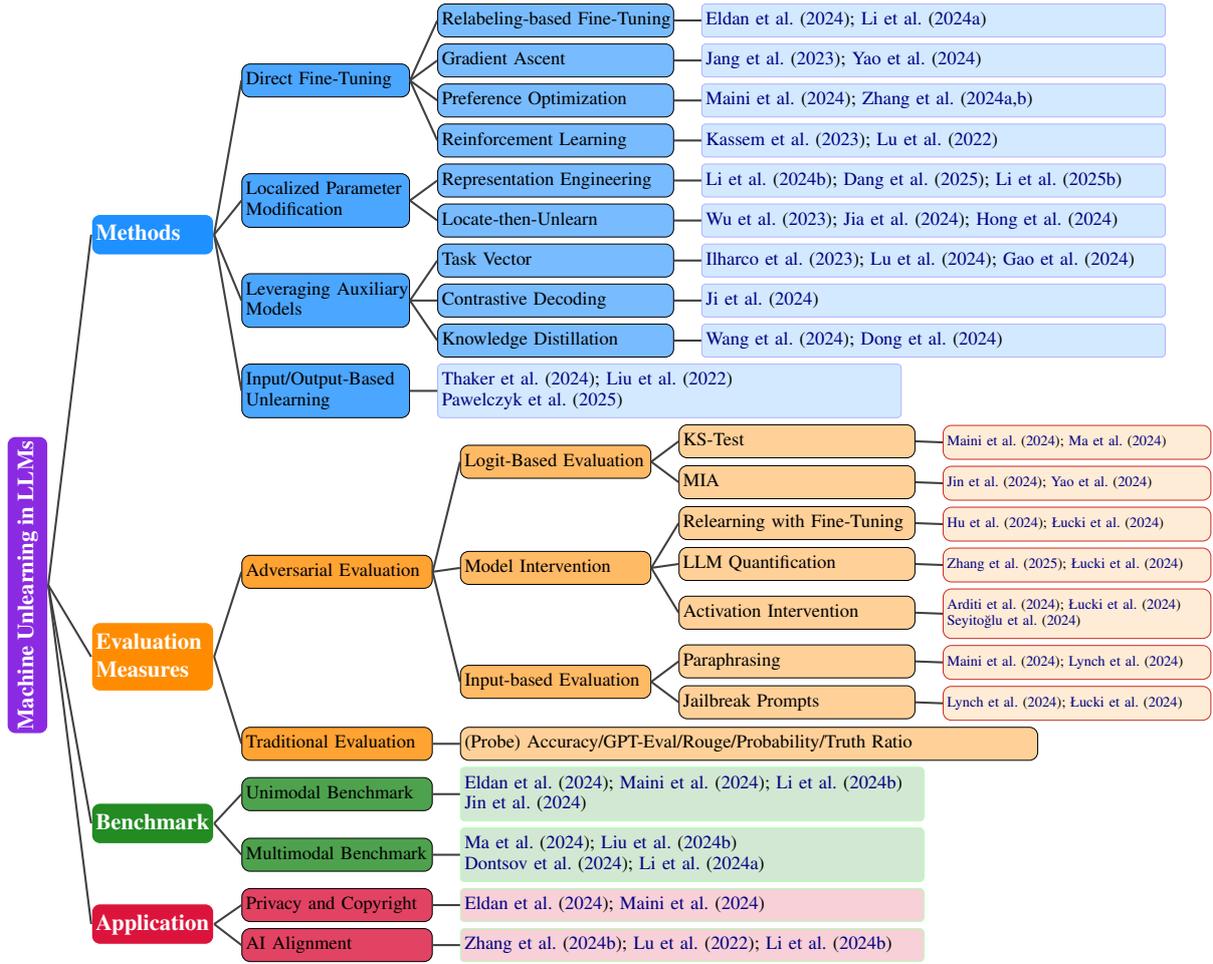
\begin{figure*}[t]
    \centering
    \resizebox{\textwidth}{!}{
        \begin{forest}
            for tree={
                grow=east,
                reversed=true,
                anchor=base west,
                parent anchor=east,
                child anchor=west,
                base=left,
                font=\small,
                rectangle,
                draw=hidden-draw,
                rounded corners,
                align=left,
                minimum width=4em,
                edge+={darkgray, line width=1pt},
                s sep=3pt,
                inner xsep=2pt,
                inner ysep=3pt,
                ver/.style={rotate=90, child anchor=north, parent anchor=south, anchor=center},
            },
            [
                Machine Unlearning in LLMs, ver, color=white, fill=rootcolor, text=white, font=\bfseries
                [
                    Methods, color=white, fill=methodcolor, text width=5.0em, text=white, font=\bfseries
                    [
                        Direct Fine-Tuning, color=black, fill=methodcolor!80, text width=7em, text=black
                        [
                            Relabeling-based Fine-Tuning, color=black, fill=methodcolor!60, text width=10.0em, text=black
                            [
                                {\citet{eldan2024whos,siu}}, detect_leaf, text width=20em, fill=methodcolor!20
                            ]
                        ]
                        [
                            Gradient Ascent, color=black, fill=methodcolor!60, text width=10.0em, text=black
                            [
                                {\citet{jang-etal-2023-knowledge,yao-etal-2024-machine}}, detect_leaf, text width=20em, fill=methodcolor!20
                            ]
                        ]
                        [
                            Preference Optimization, color=black, fill=methodcolor!60, text width=10.0em, text=black
                            [
                                {\citet{maini2024tofu,zhang2024negative,zhang2024safe}}, detect_leaf, text width=20em, fill=methodcolor!20
                            ]
                        ]
                        [
                            Reinforcement Learning, color=black, fill=methodcolor!60, text width=10.0em, text=black
                            [
                                {\citet{kassem-etal-2023-preserving,NEURIPS2022_b125999b}}, detect_leaf, text width=20em, fill=methodcolor!20
                            ]
                        ]
                    ]
                    [
                        {Localized Parameter \\ Modification}, color=black, fill=methodcolor!80, text width=7em, text=black
                        [
                            Representation Engineering, color=black, fill=methodcolor!60, text width=10.0em, text=black
                            [
                                {\citet{li2024the,huu2024effects,li2025sauce}}, detect_leaf, text width=20em, fill=methodcolor!20
                            ]
                        ]
                        [
                            Locate-then-Unlearn, color=black, fill=methodcolor!60, text width=10.0em, text=black
                            [
                                {\citet{wu-etal-2023-depn,jia2024wagle,hong2024intrinsic}}, detect_leaf, text width=20em, fill=methodcolor!20
                            ]
                        ]
                    ]
                    [
                        {Leveraging Auxiliary \\ Models}, color=black, fill=methodcolor!80, text width=7em, text=black
                        [
                            Task Vector, color=black, fill=methodcolor!60, text width=10.0em, text=black
                            [
                                {\citet{ilharco2023editing,lu2024towards,gao-etal-2024-ethos}}, detect_leaf, text width=20em, fill=methodcolor!20
                            ]
                        ]
                        [
                            Contrastive Decoding, color=black, fill=methodcolor!60, text width=10.0em, text=black
                            [
                                {\citet{ji2024reversing}}, detect_leaf, text width=20em, fill=methodcolor!20
                            ]
                        ]
                        [
                            Knowledge Distillation, color=black, fill=methodcolor!60, text width=10.0em, text=black
                            [
                                {\citet{wang2024rkld,dong2024unmemorization}}, detect_leaf, text width=20em, fill=methodcolor!20
                            ]
                        ]
                    ]
                    [
                        {Input/Output-Based\\ Unlearning}, color=black, fill=methodcolor!80, text width=7em, text=black
                        [
                            {\citet{thaker2024guardrail,liu2022continual}\\\citet{Incontextunlearning}}, detect_leaf, text width=20em, fill=methodcolor!20
                        ]
                    ]
                ]
                                [
                    Evaluation\\ Measures, color=white, fill=evalcolor, text width=5.0em, text=white, font=\bfseries
                    [
                        Adversarial Evaluation, color=black, fill=evalcolor!80, text width=8em, text=black
                        [
                            Logit-Based Evaluation, color=black, fill=evalcolor!60, text width=8.0em, text=black
                            [
                                KS-Test, color=black, fill=evalcolor!40, text width=10em, text=black
                                [
                                    {\citet{maini2024tofu,ma2024benchmarking}}, cause_leaf, text width=11.4em, fill=evalcolor!20
                                ]
                            ]
                            [
                                MIA, color=black, fill=evalcolor!40, text width=10em, text=black
                                [
                                    {\citet{jin2024rwku,yao-etal-2024-machine}}, cause_leaf, text width=11.4em, fill=evalcolor!20
                                ]
                            ]
                        ]
                        [
                            Model Intervention, color=black, fill=evalcolor!60, text width=8.0em, text=black
                            [
                                Relearning with Fine-Tuning, color=black, fill=evalcolor!40, text width=10.0em, text=black
                                [
                                    {\citet{hu2024jogging,lucki2024adversarial}}, cause_leaf, text width=11.4em, fill=evalcolor!20
                                ]
                            ]
                            [
                                LLM Quantification, color=black, fill=evalcolor!40, text width=10.0em, text=black
                                [
                                    {\citet{zhang2024does,lucki2024adversarial}}, cause_leaf, text width=11.4em, fill=evalcolor!20
                                ]
                            ]
                            [
                                Activation Intervention, color=black, fill=evalcolor!40, text width=10.0em, text=black
                                [
                                    {\citet{arditi2024refusal,lucki2024adversarial}\\ \citet{seyitouglu2024extracting}}, cause_leaf, text width=11.4em, fill=evalcolor!20
                                ]
                            ]
                        ]
                        [
                            Input-based Evaluation, color=black, fill=evalcolor!60, text width=8.0em, text=black
                            [
                                Paraphrasing, color=black, fill=evalcolor!40, text width=10.0em, text=black
                                [
                                    {\citet{maini2024tofu,lynch2024eight}}, cause_leaf, text width=11.4em, fill=evalcolor!20
                                ]
                            ]
                            [
                                Jailbreak Prompts, color=black, fill=evalcolor!40, text width=10.0em, text=black
                                [
                                    {\citet{lynch2024eight,lucki2024adversarial}}, cause_leaf, text width=11.4em, fill=evalcolor!20
                                ]
                            ]
                        ]
                    ]
                    [
                        Traditional Evaluation, color=black, fill=evalcolor!80, text width=8em, text=black
                            [   
                                (Probe) Accuracy/GPT-Eval/Rouge/Probability/Truth Ratio, color=black, fill=evalcolor!40, text width=25em, text=black
                            ]
                    ]
                ]
                [
                    Benchmark, color=white, fill=benchcolor, text width=5.0em, text=white, font=\bfseries
                    [
                        Unimodal Benchmark, color=black, fill=benchcolor!80, text width=8em, text=black
                        [
                            {\citet{eldan2024whos,maini2024tofu,li2024the}\\ \citet{jin2024rwku}}, mitigate_leaf, text width=20em, fill=benchcolor!20
                        ]
                    ]
                    [
                        Multimodal Benchmark, color=black, fill=benchcolor!80, text width=8em, text=black
                        [  
                            {\citet{ma2024benchmarking,liu2024protecting}\\ \citet{dontsov2024clear,siu}}, mitigate_leaf, text width=20em, fill=benchcolor!20
                        ]
                    ]
                ]
                [
                    Application, color=white, fill=appcolor, text width=5.0em, text=white, font=\bfseries
                    [
                        Privacy and Copyright, color=black, fill=appcolor!80, text width=8em, text=black
                        [
                            {\citet{eldan2024whos,maini2024tofu}}, mitigate_leaf, text width=20em, fill=appcolor!20
                        ]
                    ]
                    [
                        AI Alignment, color=black, fill=appcolor!80, text width=8em, text=black
                        [  
                            {\citet{zhang2024safe,NEURIPS2022_b125999b,li2024the}}, mitigate_leaf, text width=20em, fill=appcolor!20
                        ]
                    ]
                ]
            ]   
        \end{forest}
    }
\caption{The taxonomy of machine unlearning in LLMs.}
\label{fig:machine_unlearning_taxonomy}
\end{figure*}
\subsection{Taxonomy}
We present a comprehensive taxonomy of \mllm unlearning in Figure~\ref{fig:machine_unlearning_taxonomy},
outlining existing research from the perspectives of methods, evaluation measures, benchmarks, and applications. Existing methods can be categorized into four types: direct fine-tuning, localized parameter modification, leveraging auxiliary models, and input/output-based unlearning. Forgetting quality and utility preservation are critical measures for evaluating unlearning algorithms, particularly given recent discussions on whether knowledge is robustly forgotten or remains susceptible to adversarial recovery. This is often assessed through input- or logit-based evaluation, as well as model intervention techniques. Additionally, we review commonly used unimodal and multimodal  benchmarks.

\section{\mllm Unlearning Methods}
\label{sec:methods}

\subsection{Direct Fine-Tuning}
\label{sec:direct_fine_tuning}

\noindent\textbf{Relabeling-based fine-tuning} first replaces the original responses with generic or neutral substitutes, as present above. The LLM is then fine-tuned on relabeled data to reduce the effect of undesired information~\cite{jin2024rwku,eldan2024whos}.




\paragraph{Gradient ascent (GA)} \mycite{jang-etal-2023-knowledge} apply gradient ascent (GA) on the next-token loss over the forget set, equivalent to minimizing the negative log-likelihood:

\begin{equation}
\small
\mathcal{L}_{GA}(\theta) = -\mathbb{E}_{(x, y) \sim \mathcal{D}_f} \left[ -\log \left( p(y \mid x; \theta) \right) \right]. 
\end{equation}

However, GA often degrades model quality, producing uniform, low-quality outputs. To mitigate this, \mycite{liu2022continual} add a gradient descent (GD) loss on the retain set $\mathcal{D}_r$ as regularization. \mycite{yao-etal-2024-machine} instead use KL divergence to align the fine-tuned model with the original on $\mathcal{D}_r$:


\begin{equation}
\small
\textstyle
\mathcal{L}_{KL}(\theta) = \mathbb{E}_{(x, y) \sim \mathcal{D}_r} \left[ D_{KL} \left( p(y \mid x; \theta_{ori}) \, \| \, p(y \mid x; \theta) \right) \right].
\label{eq:kl_loss}
\end{equation}
These regularization techniques can also be applied in other unlearning methods to preserve the utility. 


\paragraph{Reinforcement learning (RL)}
The Quark method~\cite{NEURIPS2022_b125999b} is a pioneering approach to applying reinforcement learning for LLM unlearning. It uses a reward model and Proximal Policy Optimization (PPO)~\cite{schulman2017proximal} to reduce undesirable behaviors such as toxicity, repetition, and unwanted sentiment. The reward model assesses the output quality using task-specific measures of toxicity and sentiment. Quark alternates between collecting samples, sorting them into reward-based quantiles labeled with reward tokens, and applying language modeling loss conditioned on these tokens, with a KL-divergence penalty to stay close to the original model. \mycite{kassem-etal-2023-preserving} proposed DeMem, which leverages a negative similarity reward. This approach trains the LLMs to develop a paraphrasing policy on the forget dataset, generating dissimilar tokens that minimize memorization while preserving semantic coherence. 

\noindent \textbf{Preference optimization (PO)} was first designed to align model behavior to human-defined preferences. Specifically, it leverages pairwise comparisons or ranking data to guide the model toward producing outputs that best match desired preferences. Given a preference dataset $\mathcal{D}_{p} = \left\{ (x_i, y_{i,w}, y_{i,l}) \right\}_{i \in [n]}$, where $y_{i,w}$ and $y_{i,l}$ represent responses to input $x_i$, the preference $y_{i,w} > y_{i,l}$ is derived from human comparisons. Direct preference optimization (DPO)~\cite{rafailov2023direct} minimizes the following objective function:
\begin{equation}
\small
\textstyle
\begin{aligned}
\mathcal{L}_{\text{DPO},\beta}(\theta) = &-\frac{1}{\beta} \mathbb{E}_{\mathcal{D}_{p}} 
\left[ \log \sigma \left( 
\beta \log \frac{p(y_w \mid x; \theta)}{p(y_w \mid x; \theta_{ori})} \right. \right.\\ 
&- \left. \left. \beta \log \frac{p(y_l \mid x; \theta)}{p(y_l \mid x; \theta_{ori})} \right) \right], 
\end{aligned}
\label{eq:dpo}
\end{equation}
where $\sigma$ is the sigmoid function and $\beta$ is the inverse temperature controlling the preference strength. \mycite{maini2024tofu} pioneered the application of DPO to unlearning by framing the forget set as a preference set. The original responses are denoted as $y_{l}$, while refusal responses, such as ``\emph{I do not know the answer},'' are designated as $y_w$. This formulation guides the unlearning process by aligning the model's behavior with the preferred alternative responses. Inspired by this idea, \mycite{zhang2024negative} proposed negative preference optimization (NPO), a DPO variant that uses only negative responses from the forget set, disregarding \(y_w\) in Eq.~\eqref{eq:dpo}:
\begin{equation}
\small
\textstyle
     \mathcal{L}_{\text{DPO}, \beta}(\theta) = -\frac{2}{\beta} \mathbb{E}_{(x, y) \sim \mathcal{D}_f} \left[ \log \sigma \left( -\beta \log \frac{p(y \mid x; \theta)}{p(y \mid x; \theta_{ori})} \right) \right].
\end{equation}
\mycite{zhang2024negative} further theoretically showed that NPO converges to GA as $\beta \to 0 $ and the speed toward collapse using NPO is exponentially slower than GA. 






\subsection{Localized Parameter Modification}
\label{sec:localized_parameter_modification}

\paragraph{Representation engineering} RMU~\cite{li2024the} focuses on unlearning hazardous knowledge in LLMs by fine-tuning the lower layer $l$  to redirect internal representations of token $t$ in the forget set toward a fixed-noise vector $\mathbf{u}$: 
\begin{equation}
\small
\mathcal{L}_{\text{forget}}(\theta) = \mathbb{E}_{(x, y) \sim \mathcal{D}_f} \frac{1}{L_x} \sum_{t \in x} \| \mathcal{H}^{(l)}_{\theta}(t) - c \cdot \mathbf{u} \|_2^2,
\end{equation}
where $L_x$ is the number of tokens in $x$ and $c$ is some hyperparameter that controls activation scaling, $\mathcal{H}^{(l)}_{\theta}(t)$ denotes the internal activations of token $t$ at layer $l$. Simultaneously, it ensures that preserved knowledge remains consistent with the original model by aligning its representations, denoted as:
\begin{equation}
\small
\mathcal{L}_{\text{retain}}(\theta) = \mathbb{E}_{(x, y) \sim \mathcal{D}_r}  \frac{1}{L_x} \sum_{t \in x} \| \mathcal{H}^{(l)}_{\theta}(t) - \mathcal{H}^{(l)}_{\theta_{\text{ori}}}(t) \|_2^2. 
\end{equation}

\mycite{huu2024effects} proposed adaptive RMU, which dynamically scales the random unit vector \(\mathbf{u}\) with an adaptive coefficient \(\beta \| \mathcal{H}_{\theta_{\text{ori}}}^{(l)}(x) \|_2^2\) for improved unlearning across layers, unlike RMU's fixed-scaling coefficient \(c\).


\paragraph{Locate-then-unlearn methods} This method focuses on identifying and localizing key model components (e.g., layers or neurons) that are critical for unlearning.
DEPN~\cite{wu-etal-2023-depn} leverages privacy attribution via {gradient integration} to identify privacy-sensitive neurons. It quantifies each neuron's contribution to privacy leakage by efficiently approximating the integral of the gradient changes using a limited number of steps (e.g., $m=20$). Specifically, the privacy attribution score $\text{Att}(w_k^l)$ for a neuron $w_k^l$ at layer $l$ is computed using the following cumulative gradient integration:
\begin{equation}
\small
\text{Att}(w_k^l)=\int_0^{\beta_k^l} \frac{\partial p(y|x, \alpha_k^l)}{\partial w_k^l} d\alpha_k^l  \approx \frac{\beta_k^l}{m} \sum_{j=1}^m \frac{\partial p(y|x, \frac{j}{m} \beta_k^l)}{\partial w_k^l},
\end{equation}
where $\beta_k^l$ is the activation value of the neuron, $\alpha_k^l$ represents the modified activation value, and $p(y|x)$ is the conditional probability of the model predicting the private information. 

WAGLE~\cite{jia2024wagle} uses bi-level optimization to examine the interaction between weight adjustment and unlearning efficacy. By leveraging weight attribution, it quantifies the relationship between weight influence and the impact of forgotten or retained data on LLM outputs. The unlearning sensitivity score for weight perturbation is obtained from the forget loss $S_i = \mathcal{L}_f(\epsilon \odot \theta(\epsilon)) - \mathcal{L}_f(\theta(1))$, where \(\epsilon \odot \theta(\epsilon)\) is the weight-adjusted model, \(\epsilon\) represents weight modifications, and \(\epsilon=1\) indicates no interference. The weights \(\theta(\epsilon)\) minimize the retain loss \(\mathcal{L}_r\) to preserve utility. WAGLE uses a diagonal Hessian approximation for computational efficiency and accuracy, with the sensitivity score expressed as:
\begin{equation}
    S_i \propto [\theta]_i [\nabla \mathcal{L}_f(\theta)]_i - \frac{1}{\gamma} [\nabla \mathcal{L}_r(\theta)]_i [\nabla \mathcal{L}_f(\theta)]_i,
\end{equation}
 where \([\theta]_i\) is the \(i\)-th weight, \([\nabla \mathcal{L}_f(\theta)]_i\) and \([\nabla \mathcal{L}_r(\theta)]_i\) are the gradients of the forget and the retain losses for the \(i\)-th weight, respectively, and \(\gamma\) is the Hessian parameter.




Beyond DEPN and WAGLE, \mycite{guo2024mechanistic} introduced a mechanistic unlearning framework that combines fact lookup localization, using logistic regression probes or path patching to assess causal importance, with localized fine-tuning. Similarly, Needle~\cite{hong2024intrinsic} identifies and disrupts concept vectors in MLP layers encoding specific knowledge using vocabulary projections and causal tests. 

\subsection{Leveraging Auxiliary Models}

These methods typically fine-tune an assistant model \(\theta_{a}\) to replicate knowledge from \(\mathcal{D}_{f}\). Its outputs are then used to adjust the original model's responses, mitigating the influence of \(\mathcal{D}_{f}\) through the auxiliary model's weights or logits.

\paragraph{Contrastive decoding} 


ULD~\cite{ji2024reversing} uses an auxiliary model trained on the forget set to guide the unlearning process during decoding. It claims that the auxiliary LLM should exclusively capture the unique knowledge within the forget set while preventing the retention of any information meant to be preserved. Ideally, this results in a uniform distribution over the retain set. The optimization objective of the auxiliary model is formulated as the inverse of Eq.~\eqref{eq:definition}:
\begin{equation}
\small
\min_{\theta_a} \mathcal{L}(\theta_a) = \min_{\theta_a} \{ \mathcal{L}_f(\theta_a) - \beta \mathcal{L}_r(\theta_a) \}.
\label{eq:reverse}
\end{equation}
The retain loss \(\mathcal{L}_r(\theta_a)\) is specifically formulated as the cross-entropy with respect to the uniform distribution. To enhance the efficiency in the auxiliary model's implementation, the first \(k\) transformer layers of the original LLM are reused, along with the language model head, to map hidden representations to output logits across the entire vocabulary.


\paragraph{Knowledge distillation} uses a specialized unlearning teacher model to guide the unlearning process, providing signals to the student model to adjust the logits and to selectively forget specific information. RKLD~\cite{wang2024rkld} first identifies tokens requiring unlearning by detecting such with consistently increased logit values after fine-tuning. The formula for the unlearning teacher model is as follows:
\begin{equation}
\small
       l_{}(y|x;\theta_{ori}) - \alpha \cdot {ReLU}(l(y|x;\theta_{a}) - l(y|x;\theta_{ori}))
\end{equation}
where $l(y|x;\theta_{ori})$ and $l(y|x;\theta_{a})$ represent the logits of the original and the auxiliary model, respectively, and $\alpha$ is a hyperparameter that controls the forgetting strength, respectively.
This strategy offers more precise guidance for intentional forgetting while safeguarding other information. Moreover, \mycite{dong2024unmemorization} introduced a self-distillation method that assumes tokens like named entities or nouns contain sensitive information requiring unlearning. To identify these key tokens, they used a syntactic analysis tool for extraction. The teacher model's logits were derived by subtracting hyperparameter-controlled one-hot vectors for the key tokens from the logits of the original model.


\paragraph{Task vectors}, defined as \(\tau = \theta_{a} - \theta_{ori}\), steer the model behavior by editing the weight space, thus enabling operations such as negation and addition for applications such as unlearning and multi-task learning.
Negating task vectors effectively suppress behaviors such as mitigating toxic language generation~\cite{ilharco2023editing}. \mycite{liu-etal-2024-towards-safer} enhanced fine-tuning with modules targeting harmful knowledge: guided distortion, random disassociation, and preservation divergence. Ethos~\cite{gao-etal-2024-ethos} distinguishes general and undesired knowledge by projecting task vectors onto principal components. By negating only undesired components, Ethos minimizes collateral damage to model utility, thus achieving unlearning.



\subsection{Input/Output-based Unlearning}
Input/output-based unlearning methods offer flexibility by using prompt engineering and post-processing without modifying the model weights or the architecture. 
~\mycite{liu2024large} proposed training a prompt classifier to identify prompts within the scope of unlearning and efficiently corrupting them in the embedding space using zeroth-order optimization. \mycite{thaker2024guardrail} showed that simple guardrails like prompting and input/output filtering can effectively support unlearning independently or alongside fine-tuning. ~\mycite{Incontextunlearning} proposed in-context unlearning by constructing tailored prompts, where the labels of the data to be forgotten are flipped.

\begin{table}[]
\resizebox{0.49\textwidth}{!}{
\begin{tabular}{lccc}
\toprule
Method & \makecell{Utility \\ Preservation} & \makecell{Robust \\ Forgetting} & Efficiency \\
\midrule
Relabeling-Based Fine-Tuning  & Low                 & Medium            & High       \\
Gradient Ascent               & Low                 & High              & High       \\
Preference Optimization       & High                & Medium            & High       \\
Reinforcement Learning          & High                & Medium            & High       \\
Representation Engineering    & High                & Medium            & Medium     \\
Locate-Then-Unlearn Methods   & High                & High              & Medium     \\
Contrastive Decoding          & Medium              & Medium            & Low        \\
Task Vector                   & Low                 & Medium            & Low        \\
Knowledge Distillation        & Medium              & Medium            & Low        \\
Input/Output-based Unlearning & High                & Low               & High      \\
\bottomrule
\end{tabular}}
\caption{Comparison of various methods in terms of utility preservation, robust forgetting, efficiency.}
\label{tb:method_uti_robu_effi}
\end{table}

\paragraph{Summary of Unlearning Methodologies} Table~\ref{tb:method_uti_robu_effi} provides a comparative overview of existing unlearning methodologies, assessing their strengths and limitations across three key dimensions: utility preservation, robust forgetting, and efficiency. These methods reflect varying design principles, leading to distinct trade-offs. For example, GA excels in robustly forgetting target data but significantly compromises utility, often resulting in degraded model performance. On the other hand, methods like PO and RL strike a more favorable balance, achieving high utility with moderate forgetting and efficiency, making them appealing for scenarios requiring minimal side effects. Locate-then-unlearn techniques stand out by offering both high utility preservation and strong forgetting capabilities, although they often involve computationally intensive attribution analysis. Approaches that rely on auxiliary models, such as contrastive decoding or knowledge distillation, tend to suffer from reduced efficiency due to added model complexity. Overall, the choice of method hinges on the specific unlearning goal—whether prioritizing forgetting effectiveness, maintaining model utility, or optimizing computational cost.

\todo{ Methods such as gradient ascent, preference optimization, and reinforcement learning often struggle with utility preservation due to optimization conflicts (e.g., maximizing cross-entropy loss), requiring regularization to mitigate the performance loss. Relabeling-based fine-tuning and knowledge distillation have been reported to suffer from knowledge relearning. This issue arises when models are fine-tuned using small amounts of related or even unrelated samples, or through in-context learning~\cite{lynch2024eight}. Representation engineering methods have been reported to be more susceptible to adversarial attacks, failing to robustly erase knowledge~\cite{lucki2024adversarial}. Contrastive decoding and input/output-based unlearning are controversial because they do not truly remove knowledge from the models. Task vectors, limited by imprecision in localizing knowledge, are mainly used in AI alignment. From an efficiency standpoint, reinforcement learning and locate-then-unlearn approaches involve higher algorithmic complexity. Reinforcement learning requires training high-quality models to generate reward signals, while locate-then-unlearn methods rely on precise knowledge localization. Moreover, contrastive decoding and input/output-based methods increase the computational overhead during inference, which leads to slower generation.}

\begin{table*}[htbp]
\centering
\setlength{\tabcolsep}{3pt}
\resizebox{\textwidth}{!}{
\begin{tabular}{l|cc|c|c|c}
\hline
\multirow{2}{*}{Benchmark} & \multicolumn{2}{c|}{Data Source} & \multirow{2}{*}{\# Volume} & \multirow{2}{*}{Traditional Evaluation} & \multirow{2}{*}{Adversarial Evaluation} \\
 & Image & Text &  &  &  \\
\hline
WHP~\cite{eldan2024whos} & - & Harry Potter books & 330 & GPT-Eval/Probability & - \\
TOFU~\cite{maini2024tofu} & - & Fictitious Profiles & 4,000  & Probability/Rouge/TR & Paraphrasing \\
WMDP~\cite{li2024the} & - & Safety-related documents & 3,668 & Probe Accuracy & Jailbreak Prompts \\
RWKU~\cite{jin2024rwku} & - & Public figure knowledge & 2,4510 & Rouge & Paraphrasing, Jailbreak Prompts,MIA \\
~\cite{lucki2024adversarial} & - & WMDP & - & Probe Accuracy & Relearning,Intervention,Jailbreak Prompts,Pruning \\
~\cite{lynch2024eight} & - & WHP & - & GPT-Eval & Relearning, Jailbreak Prompts \\
\midrule
FIUBench~\cite{ma2024benchmarking} & Synthetic face images & Generated private data & 8,000 & Rouge/GPT-Eval/TR/EM & - \\
MMUBench~\cite{siu}  & Common visual concepts & Related knowledge  & 2,000 & Rouge/GPT-Eval/EM & Paraphrasing, MIA, Jailbreak Prompts  \\
MLLMU-Bench~\cite{liu2024protecting}   &  Fictitious and celebrity faces & Related profiles & 20,754 & Accuracy/Rouge & Paraphrasing  \\
CLEAR~\cite{dontsov2024clear}    & Synthetic face images & Data from TOFU  &  7,700 & TR, Probability, Rouge & Paraphrasing \\
\hline
\end{tabular}}
\caption{Overview of existing benchmarks for LLM unlearning. Some \cite{lucki2024adversarial,lynch2024eight} focus on adversarial evaluation using existing datasets. TR: Truth Ration, EM: Exact Match.}
\label{tab:benchmarks}
\end{table*}

\section{Benchmarks}
\label{sec:benchmark}

This section provides a detailed description of the commonly used benchmarks, addressing areas such as copyright, privacy, and AI alignment. Notably, WMDP and RWKU are designed for direct-unlearning, while other benchmarks are used within the fine-tuning-then-unlearning paradigm.

\subsection{Unimodal Benchmarks}

\textbf{Who is Harry Potter (WHP)}~\cite{eldan2024whos} evaluates unlearning of Harry Potter-related information using a dataset combining the original books (2.1M tokens) with synthetic content (1M tokens). Unlearning effectiveness is assessed through 330 Harry Potter-related questions scored with a GPT-4-based evaluation. 

\noindent \textbf{TOFU}~\cite{maini2024tofu} includes 200 synthetic author profiles with 20 question-answer examples each, ensuring no overlap with existing training data. The benchmark also includes 100 real-world author profiles and 117 world facts, comprehensively evaluating model utility after unlearning.

\noindent \textbf{WMDP}~\cite{li2024the} comprises 3,668 multiple-choice questions on biosecurity, cybersecurity, and chemical security, curated by experts to evaluate hazardous knowledge while excluding sensitive information. It serves as a benchmark for assessing LLMs' hazardous knowledge and unlearning methods for AI alignment.

\noindent \textbf{RWKU}~\cite{jin2024rwku} includes 200 unlearning targets centered on well-known public figures, comprising 13,131 multi-level forget probes and 11,379 neighbor probes. In addition, it incorporates a wide range of adversarial evaluation methods, including membership inference attacks (MIA), jailbreak prompts, and others.



\subsection{Multimodal Benchmarks}

\noindent \textbf{FIUBench}~\cite{ma2024benchmarking} comprises 400 synthetic faces paired with fictitious private data such as personal backgrounds, health records, criminal histories, phone numbers, occupations and incomes are randomly assigned. GPT-4o generates 20 question-answer pairs for each profile.

\noindent \textbf{MLLMU-Bench}~\cite{liu2024protecting} contains 500 fictitious profiles and 153 celebrity profiles, each with 14+ question-answer pairs evaluated in multimodal (image+text) and unimodal (text-only) settings. It features 20.7k questions, with fictitious profiles generated by GPT-4o and real celebrity profiles reviewed by experts. The test set includes 3.5k paraphrased questions and 500 modified images with pose variations. An additional utility set uses celebrity profiles.

\noindent\textbf{CLEAR}~\cite{dontsov2024clear} focuses on person unlearning, characterizing unlearning across textual and visual modalities. It generates consistent images through a comprehensive strategy and links them to the corresponding author-related questions from TOFU. It includes a total of 200 fictitious individuals linked with 3.7k visual question-answer pairs and 4k textual question-answer pairs, enabling a thorough evaluation of unimodal and multi-modal unlearning techniques.

\section{Evaluation Measures}
\label{sec:evaluation}

We group existing measures into two categories. \textit{Classical evaluation} uses standard utility metrics to assess forgetting (via performance drop on the forget set) and utility preservation (on the retain set and world knowledge). \textit{Adversarial evaluation }tests whether forgetting is robust or merely superficial suppression.

\subsection{Classical Evaluation}
To evaluate utility preservation and forgetting effectiveness, several classical metrics are commonly employed. \emph{(Probing) Accuracy} assesses whether the unlearned model maintains its world knowledge without performance degradation. \emph{GPT-Eval} uses large language models as evaluators to measure multiple dimensions beyond traditional metrics; for instance, \citet{ma2024benchmarking} employ GPT-4o Mini to score correctness, helpfulness, and relevance, producing an overall score between 0 and 1. \emph{ROUGE} quantifies the similarity between generated outputs and ground truth responses~\cite{maini2024tofu,yuan2024closer}. \emph{Probability-based} evaluation estimates the model’s confidence by computing the normalized conditional likelihood of the target output.
\emph{Truth Ratio} is specially proposed to compare the model’s likelihoods of correct versus incorrect answers for a given question and has been used in many benchmarks~\cite{maini2024tofu,ma2024benchmarking,dontsov2024clear}. Specifically, it is the ratio of the average normalized conditional probability of perturbed incorrect answers $\bar{y} \in \mathcal{A}$ to that of a paraphrased correct answer $\tilde{y}$. Both $\bar{y}$ and $\tilde{y}$ can be generated by LLMs. A lower truth ratio indicates better forgetting. It is defined as:
    \[
    R_{\text{truth}}(y) = \frac{\frac{1}{|\mathcal{A}|}\sum_{\bar{y}\in \mathcal{A}} p(\bar{y}|x;\theta_{\text{unl}})^{1/|\bar{y}|}}{p(\tilde{y}|x;\theta_{\text{unl}})^{1/|\tilde{y}|}}
    \]
On the retain and world knowledge sets, $\max(0, 1 - R_{\text{truth}}(y))$ is used to measure how well the model preserves relevant information.

\label{subsec:utility}

\subsection{Adversarial Evaluation}
\label{subsec:forget}


Recent work~\cite{lynch2024eight,ma2024benchmarking} emphasize the essence of distinguishing real forgetting from mere suppression, where suppressed knowledge can be recovered via adversarial techniques.
This motivates the summation of \textit{adversarial evaluation} to capture such subtle distinctions.

\paragraph{Input-based evaluation} assesses whether a model has truly forgotten information by modifying the input rather than probing its internal representations. \emph{Paraphrasing} involves rephrasing questions or translating them into other languages (e.g., Spanish or Russian) to test whether the model still recalls unlearned content~\cite{maini2024tofu, lynch2024eight}. \emph{Jailbreak prompts} attempt to revive forgotten knowledge by providing contextual cues or demonstrations during inference~\cite{lynch2024eight}, or by crafting adversarial inputs. These include both black-box strategies (e.g., role-playing) and white-box methods (e.g., optimized prefixes) to bypass unlearning and extract suppressed information~\cite{lucki2024adversarial}.

\paragraph{Logit-based evaluation} assesses unlearning effectiveness by analyzing changes in the model’s output distributions—typically probabilities or logits—before and after unlearning. 
\textit{Kolmogorov-Smirnov Test (KS-Test)} measures the divergence between output distributions by comparing their cumulative distribution functions (CDFs). A lower KS statistic indicates greater unlearning success. Its non-parametric nature makes it robust across various datasets and tasks~\cite{maini2024tofu,ma2024benchmarking}. Moreover, \textit{Membership Inference Attacks (MIAs)} determine whether a specific data point is part of a model's training data (member) or originates from outside the training set (non-member). Hence, it is applied to evaluate unlearning efficacy~\cite{jin2024rwku,yao-etal-2024-machine}. ~\mycite{jin2024rwku} used various MIA methods to assess the robustness of unlearning and found that many unlearning methods failed under such attacks.



\paragraph{Model intervention methods} evaluate the robustness of unlearning by directly modifying the model’s parameters, activations, or numerical precision—interventions that may inadvertently reverse unlearning and expose residual memorization. \textit{Relearning through fine-tuning} refers to the phenomenon where continued training on benign and loosely related data (e.g., "What is avian influenza?") causes the model to recover previously forgotten knowledge. This suggests that unlearning may suppress rather than fully remove the underlying representations~\cite{hu2024jogging,lucki2024adversarial}. \textit{LLM quantization} offers another perspective, where reducing the model's precision—such as converting weights to 4-bit—can increase the likelihood of forgotten content re-emerging, thereby weakening the unlearning effect~\cite{zhang2024does,lucki2024adversarial}. Similarly, \citet{lucki2024adversarial} evaluate neuron pruning as a means of assessing residual memorization. Finally, \textit{activation intervention} techniques analyze the model’s internal activations to identify and remove the so-called refusal direction—a vector derived from differences between the original and unlearned models. Suppressing this direction reduces refusal behavior and enables the model to regenerate responses that were assumed to be forgotten~\cite{arditi2024refusal,lucki2024adversarial,seyitouglu2024extracting,li2025sauce}.

\section{Future Directions}
\label{sec:challenges}



\paragraph{Theoretical frameworks are methodologically demanding.} Existing LLM unlearning methods often lack formal guarantees of effectiveness. While locate-then-unlearn approaches~\cite{wu-etal-2023-depn,jia2024wagle} enhance interpretability, they do not establish a rigorous theoretical foundation. A crucial future direction is to develop a comprehensive framework that formally defines and ensures its effectiveness. This could involve leveraging principles from information theory~\cite{jeon2024information} and other theoretical approaches to provide a more principled understanding of LLM unlearning.

\paragraph{Multimodal unlearning shows promising potential.} While numerous multimodal datasets have been introduced for multimodal unlearning, current methods remain largely confined to text-based unlearning approaches~\cite{ma2024benchmarking,liu2024protecting,dontsov2024clear}. 
Future research should prioritize the development of techniques capable of identifying and isolating modality-specific representations within MLLMs. 
Moreover, robust evaluation benchmarks are essential for assessing the effectiveness of multimodal unlearning methods in disentangling representations where knowledge is intertwined across both texts and images.
%




\paragraph{Real-world complexity is crucial for robust evaluation.} Current unlearning methods primarily focus on removing specific data points from the model, requiring explicit target data points (sequences) to be provided. However, real-world unlearning requests may differ from this assumption. A significant future direction for LLM unlearning lies in addressing more complex requests, such as entity-level unlearning, which aims to remove all knowledge related to a specific entity across diverse contexts and associations. This involves not only forgetting explicit facts but also erasing implicit or derived knowledge. \mycite{Choi2024OptOutIE} introduced datasets to evaluate the effectiveness of algorithms in entity-level unlearning tasks. Looking ahead, even more complex scenarios may emerge, such as removing all information about a specific organization, or erasing entire domains of knowledge, such as medical or criminal records.


\section{Conclusion} We provided a comprehensive survey of recent advances in LLM unlearning. We began by defining the problem and outlining the foundational settings of LLM unlearning. To offer a structured understanding, we proposed a novel taxonomy that categorizes existing research from diverse perspectives. We further explored the methodologies used to implement unlearning and evaluates the effectiveness of these approaches in achieving the desired forgetting. Finally, we examined the key challenges in the field and identified promising directions for future research, thus offering valuable insights for researchers and practitioners.

\newpage

\section*{Limitations}
This survey mainly has the following limitations:

\paragraph{No experimental benchmarks} Without original
experiments, this paper cannot offer empirical validation of the theories or concepts. This limits the paper’s ability to contribute new, verified knowledge to the field.

\paragraph{Potential omissions} We have made our best effort to compile the latest advancements. Due to the rapid development in this field, there is still a possibility that some important work may have been overlooked.


\section*{Ethics and Broader Impact}

We anticipate no significant ethical concerns in our work. As a survey of recent progress in this research area, our study does not involve experimental implementation, the use of sensitive datasets, or the employment of annotators for manual labeling.


\bibliography{custom}

\begin{thebibliography}{50}
\providecommand{\natexlab}[1]{#1}

\bibitem[{Arditi et~al.(2024)Arditi, Obeso, Syed, Paleka, Rimsky, Gurnee, and Nanda}]{arditi2024refusal}
Andy Arditi, Oscar~Balcells Obeso, Aaquib Syed, Daniel Paleka, Nina Rimsky, Wes Gurnee, and Neel Nanda. 2024.
\newblock \href {https://openreview.net/forum?id=pH3XAQME6c} {Refusal in language models is mediated by a single direction}.
\newblock In \emph{The Thirty-eighth Annual Conference on Neural Information Processing Systems}.

\bibitem[{Choi et~al.(2024)Choi, Rim, Lee, and Choo}]{Choi2024OptOutIE}
Minseok Choi, Daniel Rim, Dohyun Lee, and Jaegul Choo. 2024.
\newblock \href {https://api.semanticscholar.org/CorpusID:270562084} {Opt-out: Investigating entity-level unlearning for large language models via optimal transport}.

\bibitem[{{Council of the European Union}(2016)}]{gdpr2022}
{Council of the European Union}. 2016.
\newblock \href {https://eur-lex.europa.eu/legal-content/EN/TXT/?uri=celex%3A32016R0679} {{General data protection regulation (GDPR)}}.
\newblock Document 32016R0679.

\bibitem[{{Council of the European Union}(2024)}]{com2021laying}
{Council of the European Union}. 2024.
\newblock \href {https://eur-lex.europa.eu/eli/reg/2024/1689/oj/eng} {Laying down harmonised rules on artificial intelligence (artificial intelligence act) and amending certain union legislative acts}.
\newblock \emph{Proposal for a regulation of the European parliament and of the council}.

\bibitem[{Dang et~al.(2025)Dang, Pham, Thanh-Tung, and Inoue}]{huu2024effects}
Huu-Tien Dang, Tin Pham, Hoang Thanh-Tung, and Naoya Inoue. 2025.
\newblock On effects of steering latent representation for large language model unlearning.
\newblock In \emph{Proceedings of the AAAI Conference on Artificial Intelligence}, volume~39, pages 23733--23742.

\bibitem[{Dong et~al.(2024)Dong, Lin, Belkin, Huerta, and Vuli{\'c}}]{dong2024unmemorization}
Yijiang~River Dong, Hongzhou Lin, Mikhail Belkin, Ramon Huerta, and Ivan Vuli{\'c}. 2024.
\newblock Unmemorization in large language models via self-distillation and deliberate imagination.
\newblock \emph{arXiv preprint arXiv:2402.10052}.

\bibitem[{Dontsov et~al.(2024)Dontsov, Korzh, Zhavoronkin, Mikheev, Bobkov, Alanov, Rogov, Oseledets, and Tutubalina}]{dontsov2024clear}
Alexey Dontsov, Dmitrii Korzh, Alexey Zhavoronkin, Boris Mikheev, Denis Bobkov, Aibek Alanov, Oleg~Y Rogov, Ivan Oseledets, and Elena Tutubalina. 2024.
\newblock Clear: Character unlearning in textual and visual modalities.
\newblock \emph{arXiv preprint arXiv:2410.18057}.

\bibitem[{Eldan et~al.(2024)Eldan, Russinovich, and Russinovich}]{eldan2024whos}
Ronen Eldan, Mark Russinovich, and Mark Russinovich. 2024.
\newblock \href {https://openreview.net/forum?id=PDct7vrcvT} {Who{\textquoteright}s harry potter? approximate unlearning for {LLM}s}.

\bibitem[{Gao et~al.(2024)Gao, Niu, Tang, Avestimehr, and Annavaram}]{gao-etal-2024-ethos}
Lei Gao, Yue Niu, Tingting Tang, Salman Avestimehr, and Murali Annavaram. 2024.
\newblock \href {https://doi.org/10.18653/v1/2024.findings-naacl.132} {Ethos: Rectifying language models in orthogonal parameter space}.
\newblock In \emph{Findings of the Association for Computational Linguistics: NAACL 2024}, pages 2054--2068, Mexico City, Mexico. Association for Computational Linguistics.

\bibitem[{Guo et~al.(2024)Guo, Syed, Sheshadri, Ewart, and Dziugaite}]{guo2024mechanistic}
Phillip Guo, Aaquib Syed, Abhay Sheshadri, Aidan Ewart, and Gintare~Karolina Dziugaite. 2024.
\newblock Mechanistic unlearning: Robust knowledge unlearning and editing via mechanistic localization.
\newblock \emph{arXiv preprint arXiv:2410.12949}.

\bibitem[{Hong et~al.(2024)Hong, Yu, Yang, Ravfogel, and Geva}]{hong2024intrinsic}
Yihuai Hong, Lei Yu, Haiqin Yang, Shauli Ravfogel, and Mor Geva. 2024.
\newblock Intrinsic evaluation of unlearning using parametric knowledge traces.
\newblock \emph{arXiv preprint arXiv:2406.11614}.

\bibitem[{Hu et~al.(2024)Hu, Fu, Wu, and Smith}]{hu2024jogging}
Shengyuan Hu, Yiwei Fu, Steven Wu, and Virginia Smith. 2024.
\newblock Jogging the memory of unlearned llms through targeted relearning attacks.
\newblock In \emph{Neurips Safe Generative AI Workshop 2024}.

\bibitem[{Ilharco et~al.(2023)Ilharco, Ribeiro, Wortsman, Schmidt, Hajishirzi, and Farhadi}]{ilharco2023editing}
Gabriel Ilharco, Marco~Tulio Ribeiro, Mitchell Wortsman, Ludwig Schmidt, Hannaneh Hajishirzi, and Ali Farhadi. 2023.
\newblock \href {https://openreview.net/forum?id=6t0Kwf8-jrj} {Editing models with task arithmetic}.
\newblock In \emph{The Eleventh International Conference on Learning Representations}.

\bibitem[{Jang et~al.(2023)Jang, Yoon, Yang, Cha, Lee, Logeswaran, and Seo}]{jang-etal-2023-knowledge}
Joel Jang, Dongkeun Yoon, Sohee Yang, Sungmin Cha, Moontae Lee, Lajanugen Logeswaran, and Minjoon Seo. 2023.
\newblock \href {https://doi.org/10.18653/v1/2023.acl-long.805} {Knowledge unlearning for mitigating privacy risks in language models}.
\newblock In \emph{Proceedings of the 61st Annual Meeting of the Association for Computational Linguistics (Volume 1: Long Papers)}, pages 14389--14408, Toronto, Canada. Association for Computational Linguistics.

\bibitem[{Jeon et~al.(2024)Jeon, Jeung, Kim, No, and Choi}]{jeon2024information}
Dongjae Jeon, Wonje Jeung, Taeheon Kim, Albert No, and Jonghyun Choi. 2024.
\newblock An information theoretic metric for evaluating unlearning models.
\newblock \emph{arXiv preprint arXiv:2405.17878}.

\bibitem[{Ji et~al.(2024)Ji, Liu, Zhang, Liu, Kompella, Liu, and Chang}]{ji2024reversing}
Jiabao Ji, Yujian Liu, Yang Zhang, Gaowen Liu, Ramana~Rao Kompella, Sijia Liu, and Shiyu Chang. 2024.
\newblock Reversing the forget-retain objectives: An efficient llm unlearning framework from logit difference.
\newblock \emph{arXiv preprint arXiv:2406.08607}.

\bibitem[{Jia et~al.(2024)Jia, Liu, Zhang, Ram, Baracaldo, and Liu}]{jia2024wagle}
Jinghan Jia, Jiancheng Liu, Yihua Zhang, Parikshit Ram, Nathalie Baracaldo, and Sijia Liu. 2024.
\newblock \href {https://openreview.net/forum?id=VzOgnDJMgh} {{WAGLE}: Strategic weight attribution for effective and modular unlearning in large language models}.
\newblock In \emph{The Thirty-eighth Annual Conference on Neural Information Processing Systems}.

\bibitem[{Jin et~al.(2024)Jin, Cao, Wang, He, Yuan, Li, Chen, Liu, and Zhao}]{jin2024rwku}
Zhuoran Jin, Pengfei Cao, Chenhao Wang, Zhitao He, Hongbang Yuan, Jiachun Li, Yubo Chen, Kang Liu, and Jun Zhao. 2024.
\newblock Rwku: Benchmarking real-world knowledge unlearning for large language models.
\newblock \emph{arXiv preprint arXiv:2406.10890}.

\bibitem[{Kassem et~al.(2023)Kassem, Mahmoud, and Saad}]{kassem-etal-2023-preserving}
Aly Kassem, Omar Mahmoud, and Sherif Saad. 2023.
\newblock \href {https://doi.org/10.18653/v1/2023.emnlp-main.265} {Preserving privacy through dememorization: An unlearning technique for mitigating memorization risks in language models}.
\newblock In \emph{Proceedings of the 2023 Conference on Empirical Methods in Natural Language Processing}, pages 4360--4379, Singapore. Association for Computational Linguistics.

\bibitem[{Li et~al.(2024{\natexlab{a}})Li, Wei, Zhang, Qi, Du, Chen, Bi, and Liu}]{siu}
Jiaqi Li, Qianshan Wei, Chuanyi Zhang, Guilin Qi, Miaozeng Du, Yongrui Chen, Sheng Bi, and Fan Liu. 2024{\natexlab{a}}.
\newblock \href {http://papers.nips.cc/paper\_files/paper/2024/hash/3e53d82a1113e3d240059a9195668edc-Abstract-Conference.html} {Single image unlearning: Efficient machine unlearning in multimodal large language models}.
\newblock In \emph{Advances in Neural Information Processing Systems 38: Annual Conference on Neural Information Processing Systems 2024, NeurIPS 2024, Vancouver, BC, Canada, December 10 - 15, 2024}.

\bibitem[{Li et~al.(2024{\natexlab{b}})Li, Pan, Gopal, Yue, Berrios, Gatti, Lababidi, Wang, and Hendrycks}]{li2024the}
Nathaniel Li, Alexander Pan, Anjali Gopal, Summer Yue, Daniel Berrios, Alice Gatti, Rassin Lababidi, Alexandr Wang, and Dan Hendrycks. 2024{\natexlab{b}}.
\newblock \href {https://openreview.net/forum?id=xlr6AUDuJz} {The {WMDP} benchmark: Measuring and reducing malicious use with unlearning}.
\newblock In \emph{Forty-first International Conference on Machine Learning}.

\bibitem[{Li et~al.(2025{\natexlab{a}})Li, Geng, Chen, Song, Ma, and Karray}]{li2025internal}
Qing Li, Jiahui Geng, Zongxiong Chen, Kun Song, Lei Ma, and Fakhri Karray. 2025{\natexlab{a}}.
\newblock Internal activation revision: Safeguarding vision language models without parameter update.
\newblock \emph{arXiv preprint arXiv:2501.16378}.

\bibitem[{Li et~al.(2025{\natexlab{b}})Li, Geng, Zhu, Cai, Lyu, and Karray}]{li2025sauce}
Qing Li, Jiahui Geng, Derui Zhu, Fengyu Cai, Chenyang Lyu, and Fakhri Karray. 2025{\natexlab{b}}.
\newblock Sauce: Selective concept unlearning in vision-language models with sparse autoencoders.
\newblock \emph{arXiv preprint arXiv:2503.14530}.

\bibitem[{Li et~al.(2024{\natexlab{c}})Li, Lyu, Geng, Zhu, Panov, and Karray}]{li2024reference}
Qing Li, Chenyang Lyu, Jiahui Geng, Derui Zhu, Maxim Panov, and Fakhri Karray. 2024{\natexlab{c}}.
\newblock Reference-free hallucination detection for large vision-language models.
\newblock \emph{arXiv preprint arXiv:2408.05767}.

\bibitem[{Liu et~al.(2022)Liu, Liu, and Stone}]{liu2022continual}
Bo~Liu, Qiang Liu, and Peter Stone. 2022.
\newblock Continual learning and private unlearning.
\newblock In \emph{Conference on Lifelong Learning Agents}, pages 243--254. PMLR.

\bibitem[{Liu et~al.(2024{\natexlab{a}})Liu, Wang, Flanigan, and Liu}]{liu2024large}
Chris~Yuhao Liu, Yaxuan Wang, Jeffrey Flanigan, and Yang Liu. 2024{\natexlab{a}}.
\newblock Large language model unlearning via embedding-corrupted prompts.
\newblock \emph{arXiv preprint arXiv:2406.07933}.

\bibitem[{Liu et~al.(2025)Liu, Yao, Jia, Casper, Baracaldo, Hase, Yao, Liu, Xu, Li et~al.}]{liu2025rethinking}
Sijia Liu, Yuanshun Yao, Jinghan Jia, Stephen Casper, Nathalie Baracaldo, Peter Hase, Yuguang Yao, Chris~Yuhao Liu, Xiaojun Xu, Hang Li, and 1 others. 2025.
\newblock Rethinking machine unlearning for large language models.
\newblock \emph{Nature Machine Intelligence}, pages 1--14.

\bibitem[{Liu et~al.(2024{\natexlab{b}})Liu, Dou, Jia, Tan, Zeng, Yuan, and Jiang}]{liu2024protecting}
Zheyuan Liu, Guangyao Dou, Mengzhao Jia, Zhaoxuan Tan, Qingkai Zeng, Yongle Yuan, and Meng Jiang. 2024{\natexlab{b}}.
\newblock Protecting privacy in multimodal large language models with mllmu-bench.
\newblock \emph{arXiv preprint arXiv:2410.22108}.

\bibitem[{Liu et~al.(2024{\natexlab{c}})Liu, Dou, Tan, Tian, and Jiang}]{liu2024machine}
Zheyuan Liu, Guangyao Dou, Zhaoxuan Tan, Yijun Tian, and Meng Jiang. 2024{\natexlab{c}}.
\newblock Machine unlearning in generative ai: A survey.
\newblock \emph{arXiv preprint arXiv:2407.20516}.

\bibitem[{Liu et~al.(2024{\natexlab{d}})Liu, Dou, Tan, Tian, and Jiang}]{liu-etal-2024-towards-safer}
Zheyuan Liu, Guangyao Dou, Zhaoxuan Tan, Yijun Tian, and Meng Jiang. 2024{\natexlab{d}}.
\newblock \href {https://doi.org/10.18653/v1/2024.findings-acl.107} {Towards safer large language models through machine unlearning}.
\newblock In \emph{Findings of the Association for Computational Linguistics: ACL 2024}, pages 1817--1829, Bangkok, Thailand. Association for Computational Linguistics.

\bibitem[{Lu et~al.(2024)Lu, Isonuma, Mori, and Sakata}]{lu2024towards}
Huimin Lu, Masaru Isonuma, Junichiro Mori, and Ichiro Sakata. 2024.
\newblock Towards transfer unlearning: Empirical evidence of cross-domain bias mitigation.
\newblock \emph{arXiv preprint arXiv:2407.16951}.

\bibitem[{Lu et~al.(2022)Lu, Welleck, Hessel, and Choi}]{NEURIPS2022_b125999b}
Ximing Lu, Sean Welleck, Jack Hessel, and Yejin Choi. 2022.
\newblock \href {https://proceedings.neurips.cc/paper_files/paper/2022/file/b125999bde7e80910cbdbd323087df8f-Paper-Conference.pdf} {Quark: Controllable text generation with reinforced unlearning}.
\newblock In \emph{Advances in Neural Information Processing Systems}, volume~35, pages 27591--27609. Curran Associates, Inc.

\bibitem[{{\L}ucki et~al.(2024){\L}ucki, Wei, Huang, Henderson, Tram{\`e}r, and Rando}]{lucki2024adversarial}
Jakub {\L}ucki, Boyi Wei, Yangsibo Huang, Peter Henderson, Florian Tram{\`e}r, and Javier Rando. 2024.
\newblock An adversarial perspective on machine unlearning for ai safety.
\newblock \emph{arXiv preprint arXiv:2409.18025}.

\bibitem[{Lynch et~al.(2024)Lynch, Guo, Ewart, Casper, and Hadfield-Menell}]{lynch2024eight}
Aengus Lynch, Phillip Guo, Aidan Ewart, Stephen Casper, and Dylan Hadfield-Menell. 2024.
\newblock Eight methods to evaluate robust unlearning in llms.
\newblock \emph{arXiv preprint arXiv:2402.16835}.

\bibitem[{Ma et~al.(2024)Ma, Wang, Wang, Ma, Li, Li, Huang, Sun, Li, Choi et~al.}]{ma2024benchmarking}
Yingzi Ma, Jiongxiao Wang, Fei Wang, Siyuan Ma, Jiazhao Li, Xiujun Li, Furong Huang, Lichao Sun, Bo~Li, Yejin Choi, and 1 others. 2024.
\newblock Benchmarking vision language model unlearning via fictitious facial identity dataset.
\newblock \emph{arXiv preprint arXiv:2411.03554}.

\bibitem[{Maini et~al.(2024)Maini, Feng, Schwarzschild, Lipton, and Kolter}]{maini2024tofu}
Pratyush Maini, Zhili Feng, Avi Schwarzschild, Zachary~Chase Lipton, and J~Zico Kolter. 2024.
\newblock \href {https://openreview.net/forum?id=B41hNBoWLo} {{TOFU}: A task of fictitious unlearning for {LLM}s}.
\newblock In \emph{First Conference on Language Modeling}.

\bibitem[{Pawelczyk et~al.(2025)Pawelczyk, Neel, and Lakkaraju}]{Incontextunlearning}
Martin Pawelczyk, Seth Neel, and Himabindu Lakkaraju. 2025.
\newblock In-context unlearning: language models as few-shot unlearners.
\newblock In \emph{Proceedings of the 41st International Conference on Machine Learning}, ICML'24. JMLR.org.

\bibitem[{Rafailov et~al.(2023)Rafailov, Sharma, Mitchell, Manning, Ermon, and Finn}]{rafailov2023direct}
Rafael Rafailov, Archit Sharma, Eric Mitchell, Christopher~D Manning, Stefano Ermon, and Chelsea Finn. 2023.
\newblock \href {https://openreview.net/forum?id=HPuSIXJaa9} {Direct preference optimization: Your language model is secretly a reward model}.
\newblock In \emph{Thirty-seventh Conference on Neural Information Processing Systems}.

\bibitem[{Schulman et~al.(2017)Schulman, Wolski, Dhariwal, Radford, and Klimov}]{schulman2017proximal}
John Schulman, Filip Wolski, Prafulla Dhariwal, Alec Radford, and Oleg Klimov. 2017.
\newblock Proximal policy optimization algorithms.
\newblock \emph{arXiv preprint arXiv:1707.06347}.

\bibitem[{Seyito{\u{g}}lu et~al.(2024)Seyito{\u{g}}lu, Kuvshinov, Schwinn, and G{\"u}nnemann}]{seyitouglu2024extracting}
Atakan Seyito{\u{g}}lu, Aleksei Kuvshinov, Leo Schwinn, and Stephan G{\"u}nnemann. 2024.
\newblock Extracting unlearned information from llms with activation steering.
\newblock \emph{arXiv preprint arXiv:2411.02631}.

\bibitem[{Si et~al.(2023)Si, Zhang, Chang, Zhang, Qu, and Zhang}]{si2023knowledge}
Nianwen Si, Hao Zhang, Heyu Chang, Wenlin Zhang, Dan Qu, and Weiqiang Zhang. 2023.
\newblock Knowledge unlearning for llms: Tasks, methods, and challenges.
\newblock \emph{arXiv preprint arXiv:2311.15766}.

\bibitem[{Thaker et~al.(2024)Thaker, Maurya, Hu, Wu, and Smith}]{thaker2024guardrail}
Pratiksha Thaker, Yash Maurya, Shengyuan Hu, Zhiwei~Steven Wu, and Virginia Smith. 2024.
\newblock Guardrail baselines for unlearning in llms.
\newblock \emph{arXiv preprint arXiv:2403.03329}.

\bibitem[{Wang et~al.(2024)Wang, Zi, Sun, Zhao, and Qin}]{wang2024rkld}
Bichen Wang, Yuzhe Zi, Yixin Sun, Yanyan Zhao, and Bing Qin. 2024.
\newblock Rkld: Reverse kl-divergence-based knowledge distillation for unlearning personal information in large language models.
\newblock \emph{arXiv preprint arXiv:2406.01983}.

\bibitem[{Wu et~al.(2023)Wu, Li, Xu, Dong, Wu, Bian, and Xiong}]{wu-etal-2023-depn}
Xinwei Wu, Junzhuo Li, Minghui Xu, Weilong Dong, Shuangzhi Wu, Chao Bian, and Deyi Xiong. 2023.
\newblock \href {https://doi.org/10.18653/v1/2023.emnlp-main.174} {{DEPN}: Detecting and editing privacy neurons in pretrained language models}.
\newblock In \emph{Proceedings of the 2023 Conference on Empirical Methods in Natural Language Processing}, pages 2875--2886, Singapore. Association for Computational Linguistics.

\bibitem[{Xu(2024)}]{xu2024machine}
Yi~Xu. 2024.
\newblock Machine unlearning for traditional models and large language models: A short survey.
\newblock \emph{arXiv preprint arXiv:2404.01206}.

\bibitem[{Yao et~al.(2024)Yao, Chien, Du, Niu, Wang, Cheng, and Yue}]{yao-etal-2024-machine}
Jin Yao, Eli Chien, Minxin Du, Xinyao Niu, Tianhao Wang, Zezhou Cheng, and Xiang Yue. 2024.
\newblock \href {https://doi.org/10.18653/v1/2024.acl-long.457} {Machine unlearning of pre-trained large language models}.
\newblock In \emph{Proceedings of the 62nd Annual Meeting of the Association for Computational Linguistics (Volume 1: Long Papers)}, pages 8403--8419, Bangkok, Thailand. Association for Computational Linguistics.

\bibitem[{Yuan et~al.(2024)Yuan, Pang, Du, Chen, Zhang, and Lin}]{yuan2024closer}
Xiaojian Yuan, Tianyu Pang, Chao Du, Kejiang Chen, Weiming Zhang, and Min Lin. 2024.
\newblock A closer look at machine unlearning for large language models.
\newblock \emph{arXiv preprint arXiv:2410.08109}.

\bibitem[{Zhang et~al.(2024{\natexlab{a}})Zhang, Lin, Bai, and Mei}]{zhang2024negative}
Ruiqi Zhang, Licong Lin, Yu~Bai, and Song Mei. 2024{\natexlab{a}}.
\newblock \href {https://openreview.net/forum?id=MXLBXjQkmb} {Negative preference optimization: From catastrophic collapse to effective unlearning}.
\newblock In \emph{First Conference on Language Modeling}.

\bibitem[{Zhang et~al.(2024{\natexlab{b}})Zhang, Yang, Ke, Cui, Zheng, Wang, and Huang}]{zhang2024safe}
Zhexin Zhang, Junxiao Yang, Pei Ke, Shiyao Cui, Chujie Zheng, Hongning Wang, and Minlie Huang. 2024{\natexlab{b}}.
\newblock Safe unlearning: A surprisingly effective and generalizable solution to defend against jailbreak attacks.
\newblock \emph{arXiv preprint arXiv:2407.02855}.

\bibitem[{Zhang et~al.(2025)Zhang, Wang, Li, Wu, Tang, Liu, He, Yin, and Wang}]{zhang2024does}
Zhiwei Zhang, Fali Wang, Xiaomin Li, Zongyu Wu, Xianfeng Tang, Hui Liu, Qi~He, Wenpeng Yin, and Suhang Wang. 2025.
\newblock \href {https://openreview.net/forum?id=lHSeDYamnz} {Catastrophic failure of {LLM} unlearning via quantization}.
\newblock In \emph{The Thirteenth International Conference on Learning Representations}.

\end{thebibliography}




\end{document}